 \documentclass[pmlr,twocolumn,10pt]{jmlr} 

\usepackage{booktabs}
\usepackage{siunitx}

\usepackage[switch]{lineno}

\usepackage[utf8]{inputenc}
\usepackage{textgreek}

\theorembodyfont{\upshape}
\theoremheaderfont{\scshape}
\theorempostheader{:}
\theoremsep{\newline}

\jmlrvolume{259}
\jmlryear{2024}
\jmlrsubmitted{LEAVE UNSET}
\jmlrpublished{LEAVE UNSET}
\jmlrworkshop{Machine Learning for Health (ML4H) 2024} 

\title[Multi-Scale GNN for Alzheimer's Disease]{Multi-Scale Graph Neural Network for Alzheimer's Disease }

\author{
Anya Chauhan \Email{achauhan5@mgh.harvard.edu}\\
Ayush Noori \Email{anoori@college.harvard.edu}\\
Zhaozhi Li \Email{zl643@cornell.edu}\\
Yingnan He \Email{Yihe1@mgh.harvard.edu}\\
Michelle M Li \Email{michelleli@g.harvard.edu}\\
Marinka Zitnik \Email{marinka@hms.harvard.edu}\\
Sudeshna Das\Email{sdas5@mgh.harvard.edu}\\
}

\begin{document}

\maketitle

\begin{abstract}
Alzheimer’s disease (AD) is a complex, progressive neurodegenerative disorder characterized by extracellular A$\beta$ plaques, neurofibrillary tau tangles, glial activation, and neuronal degeneration, involving multiple cell types and pathways. Current models often overlook the cellular context of these pathways. To address this, we developed a multi-scale graph neural network (GNN) model, ALZ-PINNACLE, using brain omics data from donors spanning the entire aging to AD spectrum. ALZ-PINNACLE is based on the PINNACLE GNN framework, which learns context-aware protein, cell-type, and tissue representations within a unified latent space. ALZ-PINNACLE was trained on 14,951 proteins, 206,850 protein interactions, 7 cell types, and 48 cell subtypes or states. After pre-training, we investigated the learned embedding of \textit{APOE}, the largest genetic risk factor for AD, across different cell types. Notably, \textit{APOE} embeddings showed high similarity in microglial, neuronal, and CD8 cells, suggesting a similar role of \textit{APOE} in these cell types. Fine-tuning the model on AD risk genes revealed cell type contexts predictive of the role of \textit{APOE} in AD. Our results suggest that ALZ-PINNACLE may provide a valuable framework for uncovering novel insights into AD neurobiology. 
\end{abstract}

\paragraph*{Data and Code Availability}
All analysis and model development was performed on publicly available data GEO: \href{https://www.ncbi.nlm.nih.gov/geo/query/acc.cgi?acc=GSE268599}{GSE268599} and SRA: \href{https://ncbi.nlm.nih.gov/bioproject/916657}{PRJNA916657}.

\paragraph*{Institutional Review Board (IRB)}
Ethics approval is not required for this study.

\section{Introduction}
\label{sec:intro}

Alzheimer’s Disease (AD) is a multifaceted progressive neurodegenerative disease defined by extracellular A$\beta$ plaques, neurofibrillary tau tangles, glial activation, and neuronal degeneration. At the tissue level, multiple studies have corroborated the sequence of these pathophysiological changes— A$\beta$ triggers microglial responses, which lead to tau hyperphosphorylation and neurofibrillary tau tangles, ultimately causing synapse degeneration, neuronal loss, and cognitive decline \citep{Knopman2021}. Recent single-nucleus omics studies, such as \citet{Mathys2023} and \citet{Chen2022}, have shed light on the underlying molecular mechanisms and the cellular phenotypes associated with these pathological process. Other studies have characterized cell-cell communications in Alzheimer’s brains, such as \citet{lee2024}. However, significant gaps remain in understanding how the interplay between these molecular mechanisms and cellular communications ultimately culminate in neuronal loss. 

Indeed, AD presents a complex challenge involving dysfunction at the molecular, cellular, and brain network levels, with increasing evidence suggesting disruptions in interactions within and between these entities \citep{Rollo2023}. Thus, the complexity of the Alzheimer’s disease neurodegeneration process warrants a multiscale model. To tackle the problem, we  leveraged graph neural networks (GNNs), which are a class of machine learning models designed to operate on graph-structured data. GNNs allow flexible and effective modeling of relationships between entities in a system, and are thus a natural choice for the modeling approach \citep{Li2022}. In this study, we  developed a GNN with deep representation of cell-type specific protein-protein interaction (PPI) graphs, cell-types and subtypes and the relationships between them that is tailored to the complex processes underlying aging and AD neurodegeneration. Specifically, we adapted PINNACLE \citep{Li2024} for Alzheimer's disease data, using their publicly available code repository: \href{https://github.com/mims-harvard/PINNACLE}{https://github.com/mims-harvard/PINNACLE}.

\section{Methods}
\label{sec:methods}
In this study, we developed a GNN model for Alzheimer's disease (ALZ-PINNACLE) and fine-tuned it with data from AD genome-wide association studies (GWAS).

\subsection{Overview of ALZ-PINNACLE}
 ALZ-PINNACLE has two types of nodes: i) proteins, and ii) cell-types or subtypes, and two types of edges: i) PPI and ii) cell-type to cell-type edges, as well as a protein to cell-type attention mechanism. Our first step was to create a knowledge graph using publicly available omics data as described in the next section.
 
\subsection{Omics data processing}
\label{data-processing}
For cell-type and subtype identification and their associated gene expression, we utilized data from a previously conducted single-nucleus RNA-Sequencing (snRNA-Seq) study (GEO: GSE268599 and SRA: PRJNA916657). In the study, samples were obtained from five brain regions across 32 individuals with autopsy-confirmed findings along the normal aging-to-severe AD neuropathological continuum. NeuN+OLIG2- and NeuN-OLIG2- nuclei were enriched from each sample for sequencing. For the current study, we focused solely on nuclei derived from the Inferior Temporal Gyrus (ITG) region. Detailed procedures regarding raw data processing, including alignment, quality control, and cell-type annotation were reported in the study, for example, see \citet{Wachter2024}.

Each cell-type (\textit{e.g.}, astrocytes, microglia) was independently clustered into subclusters using the Seurat package \citep{Hao2024}. First, batch effects were corrected using the \texttt{RunHarmony} function from the Harmony package \citep{Korsunsky2019}, utilizing the top 20 principal components derived from the 2,000 most variable genes. Subclusters were then identified using Seurat’s \texttt{FindClusters} function at a resolution of 0.3, and each subcluster was annotated based on its enriched functional attributes. Immune cell subcluster labels were annotated by reference mapping to annotated PBMCs using Azimuth \citep{Hao2021}. We refer to these subclusters as cell ``subtypes'' or states. 

For cell-type subcluster-specific gene expression, we performed differential expression of each cell-type subcluster relative to other subclusters (\textit{e.g.}, disease associated microglia vs. other microglia subclusters) using \texttt{FindAllMarkers} in Seurat (v4.3.0). Differentially expressed genes (DEGs) that pass the average fold-change (FC) cut off (FC $\ge$ 1.2 for up-regulated genes and FC $\leq$ 0.8 for the down-regulated genes), with an adjusted $\textit{p}$-value $\le$ 0.05 and expressed in greater than 5\% of the cells were included.

For the edges between cell subtypes (subclusters), we
also used the annotated snRNAseq data from the ITG region. We constructed comprehensive ligand-receptor interactions between cell subclusters using LIANA \citep{Dimitrov2022}, which utilizes multiple cell-cell communication methods (\textit{e.g.}, CellPhoneDB, CellChat) to provide an overall ranking. The ligand-receptor interaction database used in this analysis was sourced from the LIANA ``consensus" subset. We performed the interaction analysis using the \texttt{rank\_aggregate} function from the LIANA package with default settings, and interactions with a specific rank $\le$ 0.05 were selected for graph construction. 
\begin{table*}[hbtp]
\small 
\floatconts
  {tab:table1}%
  {\caption{Data Summary}\vspace{-0.2cm}}%
  {\begin{tabular}{|>{\raggedright\arraybackslash}m{2cm} |>{\raggedright\arraybackslash}m{4cm} |>{\raggedright\arraybackslash}m{3.5cm} |>{\raggedright\arraybackslash}m{6cm}|} \hline  
      & Global PPI & Metagraph & Contextual PPI \\ \hline 
      PINNACLE &
      Proteins: 15,461\newline
      Edges: 207,641\newline
      Graph Density: 0.001737 & 
      Cells, Tissues\newline
      Cells: 156\newline
      Tissues: 24 &
      Unique Proteins: 13,643\newline 
      Protein Representations: 394,760 \\ \hline
      ALZ-PINNACLE & 
      Proteins: 14,951\newline
      Edges: 206,850\newline
      Graph Density: 0.001851 & 
      Cell-types \& Subtypes \newline
      Cell Subtypes: 48\newline
      Cell-types: 7 & 
      Unique Proteins: 14,951\newline
      Protein Representations: 365,056\newline
      Graph Density: 0.0007-0.007 \\ \hline
 \end{tabular}} 
\normalsize 
\end{table*}

For the fine-tuning with AD risk genes, positive AD risk genes were downloaded from \href{https://adsp.niagads.org/gvc-top-hits-list}{https://adsp.niagads.org/gvc-top-hits-list}. This list contains either the nearest or reported gene from AD loci with genetic evidence compiled by the ADSP Gene Verification Committee. Additionally, we incorporated new AD risk loci genes reported by a GWAS study involving 1,126,563 individuals \citep{wightman2021genome}. Negative AD risk genes were collected from a co-localization analysis by \citet{bryois2022cell}. These genes were neither eQTLs nor GWAS signals at the AD loci, identified using the criteria \texttt{PP.H0.abf} $\ge$ 0.5 and \texttt{PP.H4.abf} $\le$ 0.5.


\subsection{Pre-training}
Similar to PINNACLE \citep{Li2024}, ALZ-PINNACLE was trained using the cell-type identity and graph connectivity of cell-type-specific protein interaction (PPI) networks and a metagraph of edges between cell-types and subtypes. Pre-training was formulated a self-supervised link prediction task, \textit{e.g.}, predicting the cell subtype(s) in which a given protein was activated. In this process, a subset of edges were masked, and the model was tasked with predicting both the existence of true edges and the absence of negatively sampled false edges. Protein-protein edges were split into train (80\%), validation (10\%), and test (10\%) sets, while metagraph edges remained unsplit. For link prediction, false edges (\textit{i.e.}, negative samples) were randomly generated, with a 1:1 ratio of positive to negative edges. We trained ALZ-PINNACLE on a GPU workstation with two NVIDIA Quadro RTX 8000 GPUs. ALZ-PINNACLE required approximately six hours to pre-train for 125 epochs.
\subsection{Fine-tuning}
After pretraining, ALZ-PINNACLE was fine-tuned to predict whether the gene encoding the protein is an AD risk gene. By feeding ALZ-PINNACLE-generated embeddings of genes/proteins into a multi-layer perceptron (MLP), the model outputs a score between 0 and 1, indicating the likelihood of the gene being associated with AD risk. The cell-type-specific representation of each protein is scored independently, allowing identification of the most predictive cell-types (\textit{i.e.}, the most relevant cell types).
\section{Results}

\subsection{Alzheimer knowledge graph statistics}
The number of global PPI proteins, edges, and graph density of PINNACLE and ALZ-PINNACLE are presented in   \tableref{tab:table1}. PINNACLE had a metagraph of 156 cell-types and 24 tissues whereas ALZ-PINNACLE has 48 unique cell subtypes (\textit{e.g.}, homeostatic astrocyes) and 7 cell-types (excitatory neurons, inhibitory neurons, astrocytes, microglia, endothelial cells, immune, and brain cells).

\subsection{Pre-training results of ALZ-PINNACLE}
Performance metrics -- including AUROC (AUC), AP, accuracy (ACC), and F1 scores -- of the held-out test are shown in (\appendixref{apd:appdx1}). 

Next, we compared the similarity of ALZ-PINNACLE protein representations within a specific cell-type and compared them to their embeddings in other cell-types. (\appendixref{apd:appdx2}). We selected the proteins that had higher embedding similarities  within a cell-type as compared to their similarity in other cell-types. These proteins that have highly cell-type-specific functions are likely to be the product of marker genes, and such an analysis can be helpful for identifying cell-type-specific protein functions.


To investigate the cell-type-specific role of \textit{APOE}, the largest genetic risk factor of AD, we computed pairwise cosine similarity of \textit{APOE} cell-type-specific embeddings across the 48 different cell subtypes (\appendixref{apd:appdx3}). The  \textit{APOE} embedding similarities suggest that the role of \textit{APOE} is similar in microglia, neuronal, and CD8+ T cells (see first 4 columns). Microglia subtype \texttt{mic 5} is enriched in neurogenesis genes; \texttt{L4 IT} are layer 4 intratelencephalic (IT) neurons, and parvalbumin fast-spiking (\textit{i.e.}, \texttt{Pvalb}) neurons are a neuronal subtype essential for maintaining the excitation-inhibition balance in the brain.  

Next we investigated the cosine similarities of the cell-type embeddings   (\appendixref{apd:appdx4}). The heatmap shows that several neuronal, microglia, and astrocyte cell-subtypes are in the same neighborhood (\textit{i.e.}, there may be cell-cell interactions among them). Further investigation of the molecular phenotypes of these subclusters may yield cell-subtype communities that are resilient or vulnerable to AD. 

\subsection{Fine-tuning results}
After pre-training ALZ-PINNACLE, we fine-tuned the model on AD GWAS data (78 positive samples and 65 negative samples). The fine-tuning performance is shown in \figureref{fig:fig3}.

\begin{figure}[htbp]
\floatconts
  {fig:fig3}
  {\vspace{-1cm}\caption{Fine-tuning performance of the top 10 cell-subtypes (AP=Average Precision).}}  
  {\includegraphics[width=0.9\linewidth]{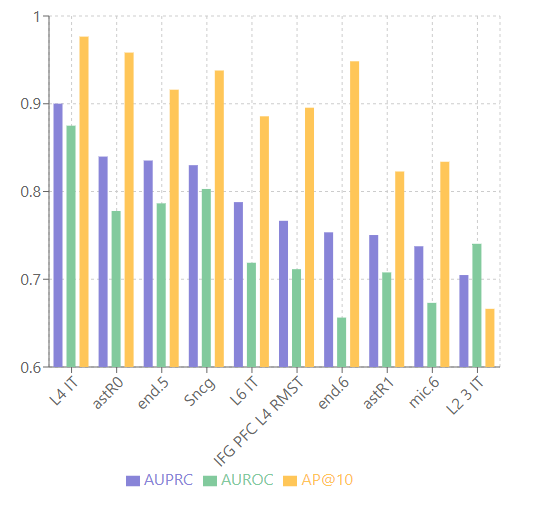}}
\end{figure}

Using the fine-tuned model, we investigated which cell type contexts were most predictive of the role of \textit{APOE} in AD (\figureref{fig:fig4}). Astrocyte, endothelial, and neuronal subclusters were predicted to have the most important roles of \textit{APOE} in AD.

\begin{figure}[htbp]
\floatconts
  {fig:fig4}
  {\vspace{-0.9cm}\caption{Most predictive cell type contexts of the role of \textit{APOE} in AD.}}  
  {\includegraphics[width=0.9\linewidth]{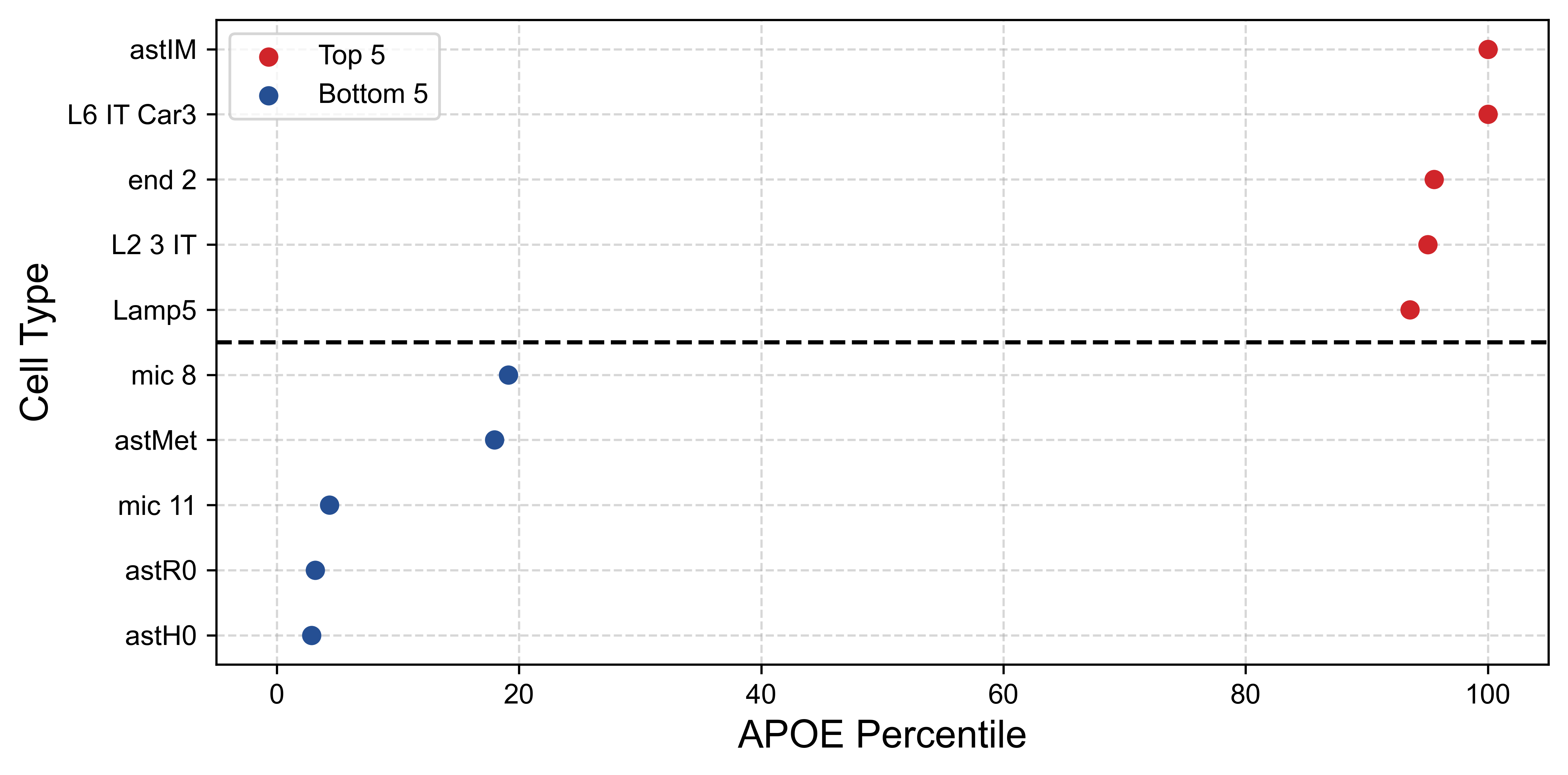}}
\end{figure}

We evaluated ALZ-PINNACLE’s performance against two context-free baseline models: a random walk algorithm applied on the reference protein interaction network and BIONIC, a graph convolutional neural network designed for multimodal network integration \citep{forster2022bionic}. BIONIC was trained for 50 epochs with the following hyperparameters: $\lambda = 0.95$, batch size $= 16$, and learning rate $= 1 \times 10^{-4}$. Encouragingly, we found ALZ-PINNACLE’s protein representations outperformed the baseline models (\tableref{tab:table2}). A head-to-head comparison between PINNACLE and ALZ-PINNACLE was not possible as they are trained on different knowledge graphs and learn representations of different cell types and cell subtypes. 


\begin{table}[htbp]
\small
\floatconts
  {tab:table2}%
  {\caption{Percentage of cell subtypes in which ALZ-PINNACLE outperformed baseline models.}}%
  {\begin{tabular}{|c|c|c|}
    \hline
    \textbf{Metric} & \shortstack{\textbf{Comparison w/} \\ \textbf{Random Walk}} & \shortstack{\textbf{Comparison w/} \\ \textbf{BIONIC}} \\ \hline
    AP@5 & 70.83\% & 60.42\% \\ \hline
    AP@10 & 64.58\% & 50.00 \\ \hline
    AUPRC & 72.92\% & 50.00 \\ \hline
    AUROC & 31.25\% & 41.67 \\ \hline
    Precision@5 & 62.50\% & 8.33\% \\ \hline
    Precision@10 & 29.17\% & 10.42\% \\ \hline
    Recall@5 & 77.08\% & 14.58\% \\ \hline
    Recall@10 & 41.67\% & 18.75\% \\ \hline
  \end{tabular}} 
\normalsize
\end{table}

\section{Conclusion and Future Directions}
Our multiscale graph neural network model, ALZ-PINNACLE, successfully integrates brain omics data across the aging to AD continuum, offering context-aware protein and cell-type representations. The model's ability to capture meaningful embeddings, such as \textit{APOE} similarity across microglial, neuronal, and CD8+ T cells, and identify cell-specific roles of AD risk genes demonstrates its potential to bridge molecular, cellular, and tissue-level insights in AD research. These findings highlight ALZ-PINNACLE potential for advancing our understanding of AD neurodegeneration.  Our model has few limitations: a single dataset is used for model developmnent, and spatial and temporal relationships are not represented. Moreover astrocyte proteins are not well represented.

In the future, these results need to be validated with follow-on experiments. ALZ-PINNACLE can be further refined by incorporating additional datasets, such as spatial transcriptomics and proteomics, to model the spatial relationship of cells to each other and to A$\beta$ plaques, neurofibrillary tau tangles. Expanding the model to explore longitudinal changes in protein interactions and cellular states during disease progression may yield deeper insights into the temporal dynamics of AD. Finally, ALZ-PINNACLE may be used for \textit{in silico} gene knock-out experiments to enable therapeutic target discovery.

\clearpage
\section{Citations and Bibliography}


\bibliography{Alz-GNN}

\begin{thebibliography}{15}
\providecommand{\natexlab}[1]{#1}
\providecommand{\url}[1]{\texttt{#1}}
\expandafter\ifx\csname urlstyle\endcsname\relax
  \providecommand{\doi}[1]{doi: #1}\else
  \providecommand{\doi}{doi: \begingroup \urlstyle{rm}\Url}\fi

\bibitem[Bryois et~al.(2022)Bryois, Calini, Macnair, Foo, Urich, Ortmann, Iglesias, Selvaraj, Nutma, Marzin, et~al.]{bryois2022cell}
Julien Bryois, Daniela Calini, Will Macnair, Lynette Foo, Eduard Urich, Ward Ortmann, Victor~Alejandro Iglesias, Suresh Selvaraj, Erik Nutma, Manuel Marzin, et~al.
\newblock Cell-type-specific cis-eqtls in eight human brain cell types identify novel risk genes for psychiatric and neurological disorders.
\newblock \emph{Nature neuroscience}, 25\penalty0 (8):\penalty0 1104--1112, 2022.

\bibitem[Chen et~al.(2022)Chen, Chang, Li, Acosta, Li, Guo, and et~al.]{Chen2022}
S.~Chen, Y.~Chang, L.~Li, D.~Acosta, Y.~Li, Q.~Guo, and et~al.
\newblock Spatially resolved transcriptomics reveals genes associated with the vulnerability of middle temporal gyrus in alzheimer's disease.
\newblock \emph{Acta Neuropathologica Communications}, 10\penalty0 (1):\penalty0 188, 2022.
\newblock \doi{10.1186/s40478-022-01494-6}.

\bibitem[Dimitrov et~al.(2022)]{Dimitrov2022}
D.~Dimitrov et~al.
\newblock Comparison of methods and resources for cell-cell communication inference from single-cell rna-seq data.
\newblock \emph{Nature Communications}, 13:\penalty0 3224, 2022.

\bibitem[Forster et~al.(2022)Forster, Li, Yashiroda, and et~al.]{forster2022bionic}
Daniel~T Forster, Shuxi~C Li, Yuki Yashiroda, and et~al.
\newblock Bionic: biological network integration using convolutions.
\newblock \emph{Nature Methods}, 19:\penalty0 1250--1261, 2022.

\bibitem[Hao et~al.(2021)]{Hao2021}
Y.~Hao et~al.
\newblock Integrated analysis of multimodal single-cell data.
\newblock \emph{Cell}, 184:\penalty0 3573--3587.e3529, 2021.

\bibitem[Hao et~al.(2024)]{Hao2024}
Y.~Hao et~al.
\newblock Dictionary learning for integrative, multimodal and scalable single-cell analysis.
\newblock \emph{Nature Biotechnology}, 42:\penalty0 293--304, 2024.

\bibitem[Knopman et~al.(2021)Knopman, Amieva, Petersen, Chetelat, Holtzman, Hyman, and et~al.]{Knopman2021}
D.~S. Knopman, H.~Amieva, R.~C. Petersen, G.~Chetelat, D.~M. Holtzman, B.~T. Hyman, and et~al.
\newblock Alzheimer disease.
\newblock \emph{Nature Reviews Disease Primers}, 7\penalty0 (1):\penalty0 33, 2021.
\newblock \doi{10.1038/s41572-021-00269-y}.

\bibitem[Korsunsky et~al.(2019)]{Korsunsky2019}
I.~Korsunsky et~al.
\newblock Fast, sensitive and accurate integration of single-cell data with harmony.
\newblock \emph{Nature Methods}, 16:\penalty0 1289--1296, 2019.

\bibitem[Lee et~al.(2024)Lee, Riffle, Xiong, Momtaz, Lei, Pariser, Sikdar, Hwang, Duan, and Zhang]{lee2024}
Che~Yu Lee, Dylan Riffle, Yifeng Xiong, Nadia Momtaz, Yutong Lei, Joseph~M Pariser, Diptanshu Sikdar, Ahyeon Hwang, Ziheng Duan, and Jing Zhang.
\newblock Characterizing dysregulations via cell-cell communications in alzheimer’s brains using single-cell transcriptomes.
\newblock \emph{BMC Neuroscience}, 25\penalty0 (1):\penalty0 24, 2024.

\bibitem[Li et~al.(2022)Li, Huang, and Zitnik]{Li2022}
M.~M. Li, K.~Huang, and M.~Zitnik.
\newblock Graph representation learning in biomedicine and healthcare.
\newblock \emph{Nature Biomedical Engineering}, 6\penalty0 (12):\penalty0 1353--1369, 2022.
\newblock \doi{10.1038/s41551-022-00942-x}.

\bibitem[Li et~al.(2024)Li, Huang, Sumathipala, Liang, Valdeolivas, Ananthakrishnan, Liao, Marbach, and Zitnik]{Li2024}
Michelle~M Li, Yepeng Huang, Marissa Sumathipala, Man~Qing Liang, Alberto Valdeolivas, Ashwin~N Ananthakrishnan, Katherine Liao, Daniel Marbach, and Marinka Zitnik.
\newblock Contextual ai models for single-cell protein biology.
\newblock \emph{Nature Methods}, pages 1--12, 2024.

\bibitem[Mathys et~al.(2023)Mathys, Peng, Boix, Victor, Leary, Babu, and et~al.]{Mathys2023}
H.~Mathys, Z.~Peng, C.~A. Boix, M.~B. Victor, N.~Leary, S.~Babu, and et~al.
\newblock Single-cell atlas reveals correlates of high cognitive function, dementia, and resilience to alzheimer's disease pathology.
\newblock \emph{Cell}, 186\penalty0 (20):\penalty0 4365--4385.e27, 2023.
\newblock \doi{10.1016/j.cell.2023.08.039}.

\bibitem[Rollo et~al.(2023)Rollo, Crawford, and Hardy]{Rollo2023}
J.~Rollo, J.~Crawford, and J.~Hardy.
\newblock A dynamical systems approach for multiscale synthesis of alzheimer's pathogenesis.
\newblock \emph{Neuron}, 111\penalty0 (14):\penalty0 2126--2139, 2023.
\newblock \doi{10.1016/j.neuron.2023.04.018}.

\bibitem[Wachter et~al.(2024)]{Wachter2024}
A.~Wachter et~al.
\newblock Landscape of brain myeloid cell transcriptome along the spatiotemporal progression of alzheimer’s disease reveals distinct sequential responses to aβ and tau.
\newblock \emph{Acta Neuropathologica}, 147:\penalty0 65, 2024.

\bibitem[Wightman et~al.(2021)Wightman, Jansen, Savage, Shadrin, Bahrami, Holland, Rongve, B{\o}rte, Winsvold, Drange, et~al.]{wightman2021genome}
Douglas~P Wightman, Iris~E Jansen, Jeanne~E Savage, Alexey~A Shadrin, Shahram Bahrami, Dominic Holland, Arvid Rongve, Sigrid B{\o}rte, Bendik~S Winsvold, Ole~Kristian Drange, et~al.
\newblock A genome-wide association study with 1,126,563 individuals identifies new risk loci for alzheimer’s disease.
\newblock \emph{Nature genetics}, 53\penalty0 (9):\penalty0 1276--1282, 2021.

\end{thebibliography}

\onecolumn
\appendix

\section{Link prediction pre-training performance for cell-type and subtype PPIs in the held out test set.}\label{apd:appdx1}

\begin{figure*}[htbp]
\floatconts
  {fig:appendixa}%
  {}
  {\includegraphics[width=0.9\linewidth]{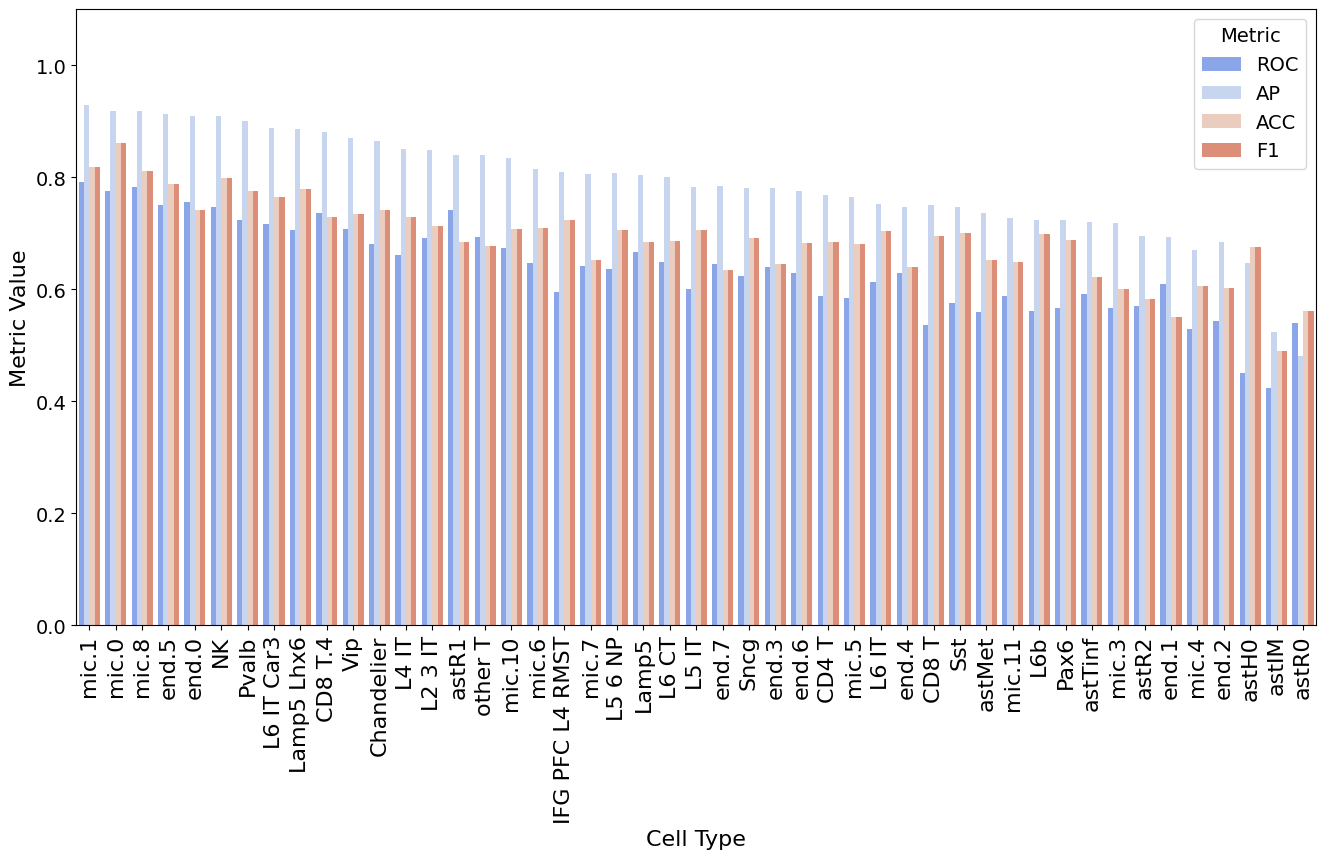}}
\end{figure*}

\newpage
\section{Similarity of protein representations.}\label{apd:appdx2}
\begin{figure*}[htbp]
\floatconts
  {fig:appendixb}%
  {}
  {\includegraphics[width=0.9\linewidth]{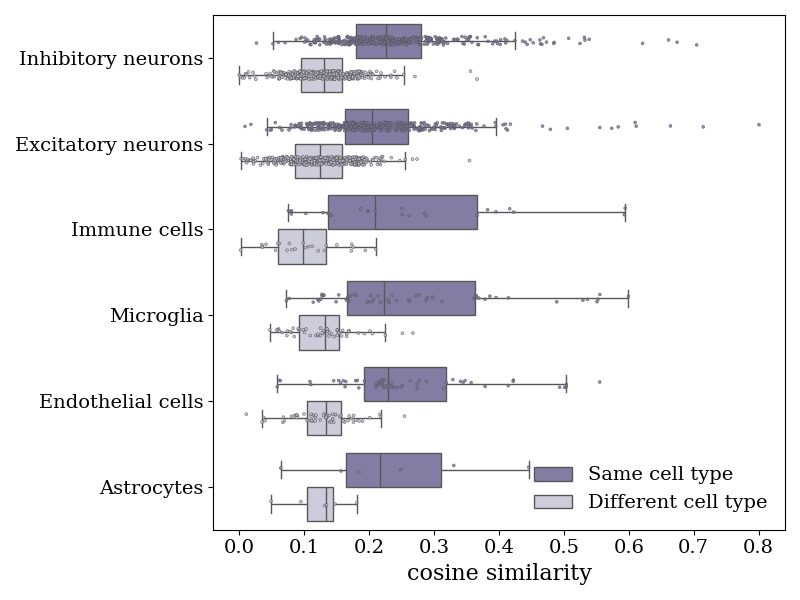}}
\end{figure*}

\newpage

\section{Cosine similarity of cell-type specific \textit{APOE} embeddings.}\label{apd:appdx3}
\begin{figure*}[htbp]
\floatconts
  {fig:appendixc}%
  {}
  {\includegraphics[width=1\linewidth]{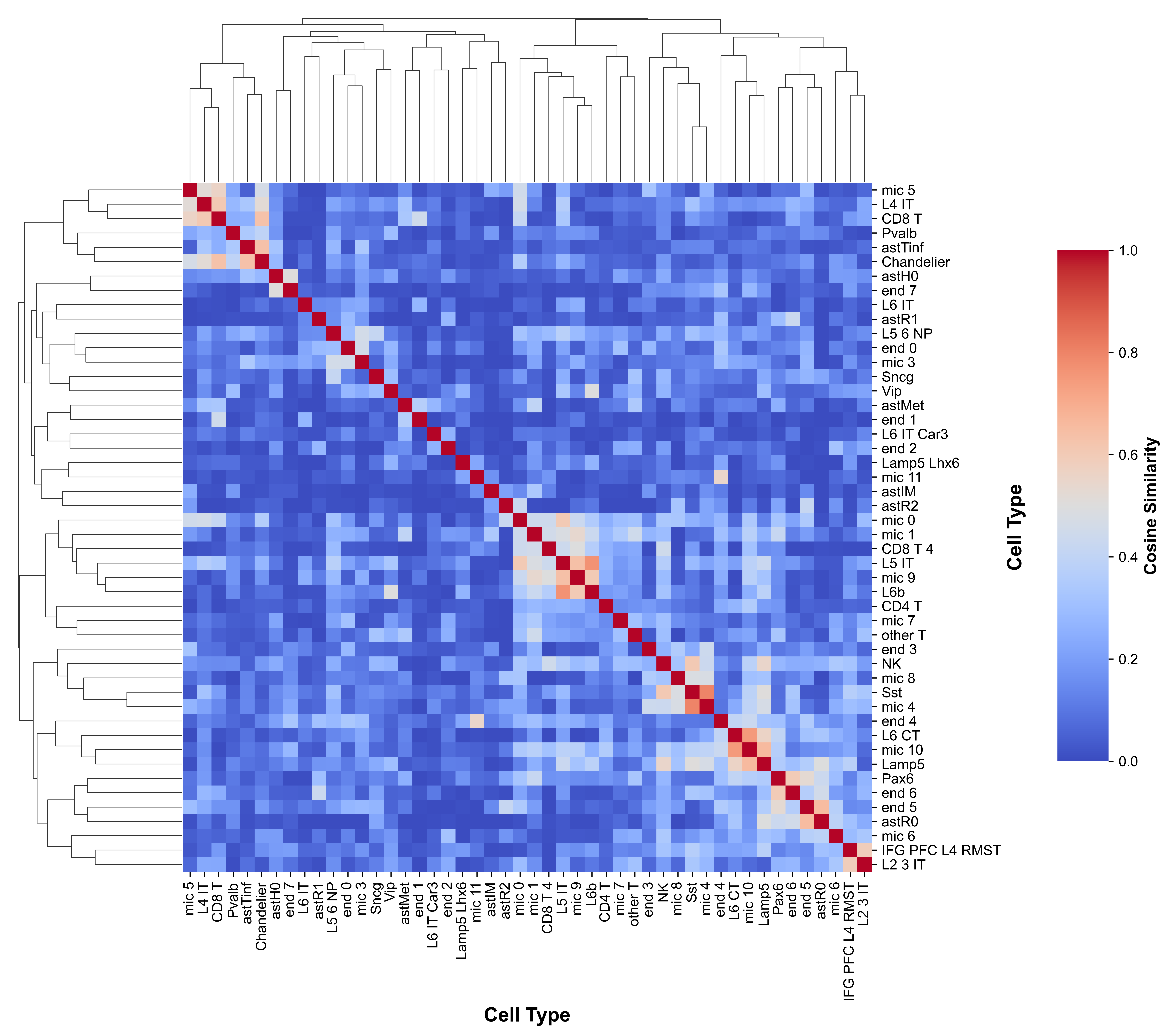}}
\end{figure*}

\newpage
\section{Cosine similarity of cell-type embeddings.}\label{apd:appdx4}
\begin{figure*}[htbp]
\floatconts
  {fig:appendixd}%
  {}
  {\includegraphics[width=1\linewidth]{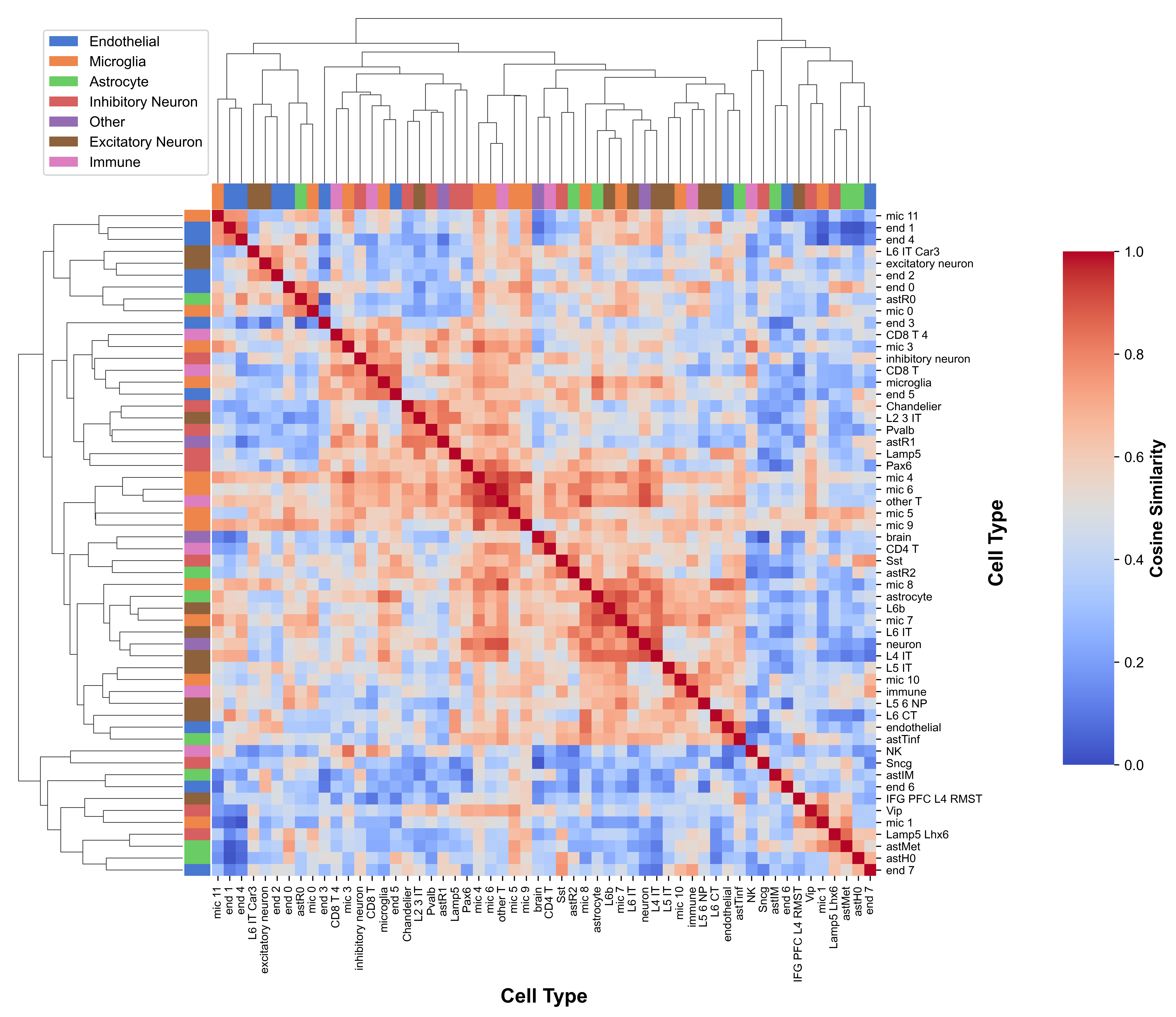}}
\end{figure*}

\end{document}